\documentclass[12pt]{article}

\usepackage{scicite}
\usepackage{times}
\usepackage{graphics} 
\usepackage{epsfig} 
\usepackage{mathptmx} 
\usepackage{times} 
\usepackage{amsmath} 
\usepackage{amssymb}  
\usepackage[utf8]{inputenc}

\usepackage{tabularx}
\usepackage{booktabs}
\usepackage{multirow}

\usepackage{algorithm}
\usepackage{algorithmic}

\usepackage{siunitx}
\usepackage{xcolor}
\usepackage[normalem]{ulem}
\usepackage{comment}

\newcommand{\refig}[1] {Fig.~\ref{#1}}

\newenvironment{sciabstract}{%
\begin{quote} \bf}
{\end{quote}}

\title{Robust bipedal locomotion on flowable slopes via foot-driven terrain manipulation}

\author{%
\hspace*{-0.15in}Deniz Kerimoglu$^{1\ast}$, Junnosuke Kamohara$^{1}$, Jiyeon Maeng$^{2}$,\\
\hspace*{-0.15in}Ziwon Yoon$^{1}$, Seth Hutchinson$^{3}$, Ye Zhao$^{1}$, and Daniel I. Goldman$^{2}$\\[0.8em]
\hspace*{-0.15in}{\normalsize $^{1}$Department of Mechanical Engineering, Georgia Institute of Technology,}\\
\hspace*{-0.15in}{\normalsize Atlanta, GA 30332, USA}\\
\hspace*{-0.15in}{\normalsize $^{2}$School of Physics, Georgia Institute of Technology, Atlanta, GA 30332, USA}\\
\hspace*{-0.15in}{\normalsize $^{3}$Department of Electrical and Computer Engineering, Northeastern University,}\\
\hspace*{-0.15in}{\normalsize Boston, MA 02115, USA}\\
\hspace*{-0.15in}{\normalsize $^{\ast}$Corresponding author; E-mail: dkerimoglu6@gatech.edu}%
}

\date{}

\begin{document} 
\baselineskip24pt
\maketitle 

\begin{sciabstract}
Bipedal robots are challenging to control because they operate close to instability, where small variations in foot-terrain contact can rapidly destabilize locomotion. On rigid terrain, bipedal robots mitigate this fragility by using well-established contact mechanics and control strategies. On flowable surfaces such as granular slopes, foot contact can induce large surface deformations and solid–fluid–like transitions, coupling terrain effects with robot dynamics, leading to underperformance or failure. This is partly due to the lack of reliable methods to represent the dynamics of flowable terrain, making it difficult to account for terrain effects in locomotion design. Here, we investigate how controlling terrain response can improve bipedal locomotion on granular slopes by studying the terradynamics of ``cleated" feet, thin plates emanating from the foot soles. Systematic studies of a small-scale (1.4 kg) robophysical biped reveal that cleats with sparse and dense spacing lead to excessive terrain yielding and resistance, respectively, degrading performance and leading to failure. An intermediate cleat spacing distributes interaction forces to maintain substrate stresses near (or below) the yield threshold, enabling walking on granular slopes up to 30°. Guided by these principles, we design a foot that actively adjusts cleat depth and accommodates both rigid and granular terrain. We also demonstrate that the principles of effective foot-terrain interaction translate to a larger (15 kg) autonomous biped. Our study presents an alternative to conventional ``body-centric" robot control approaches, which regulate terrain-induced effects through body motion, by instead regulating terrain interactions through ``limb-centric" approach. https://anonymous.4open.science/w/flowable-slope-locomotion-biped/
\end{sciabstract}

\section{INTRODUCTION}
\begin{figure}[ht]
	\centering
	\includegraphics[width=.99\columnwidth]{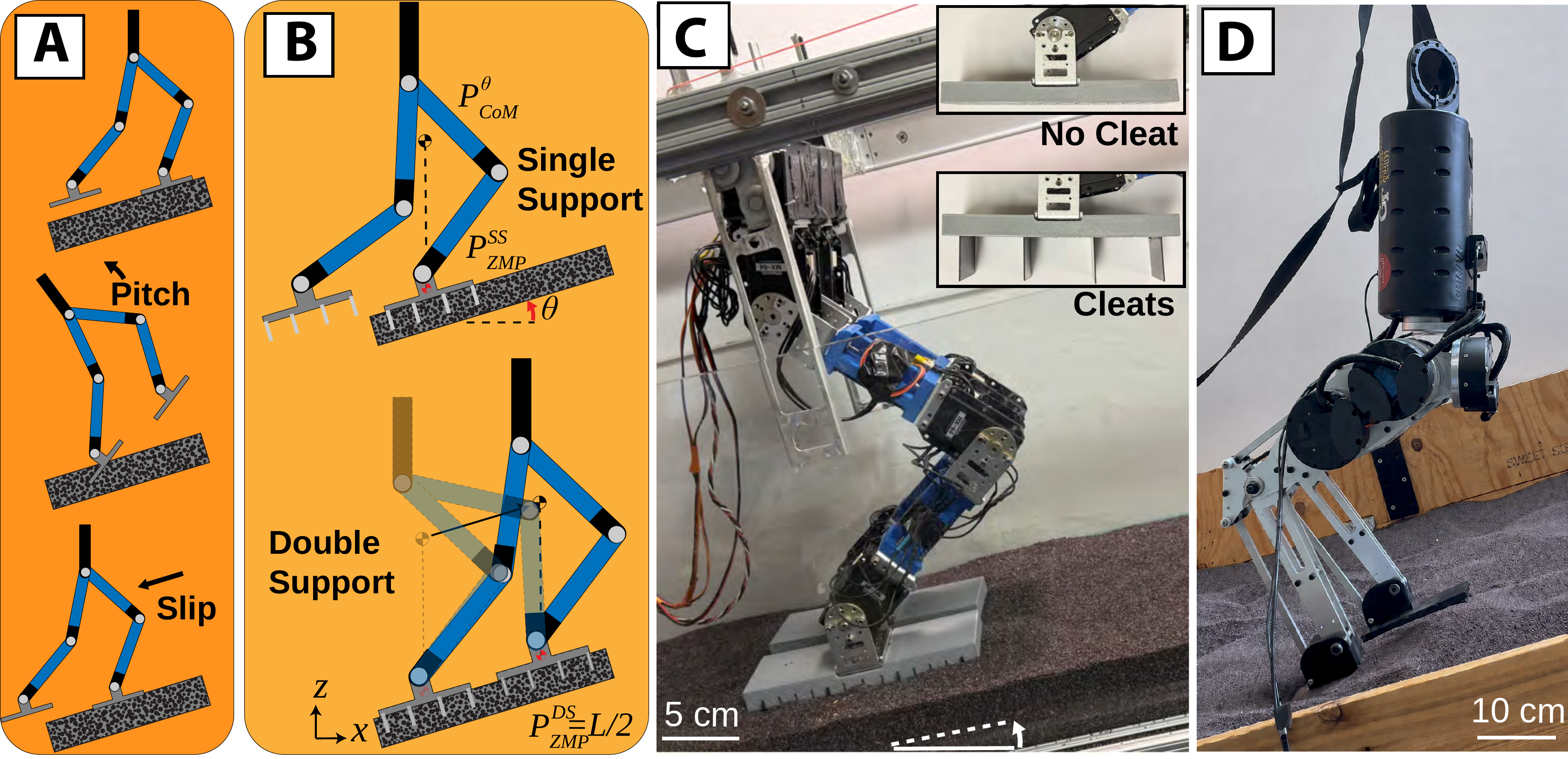}
	\caption{\textbf{Overview of cleat-foot morphology with bipedal robots on granular slopes.} A) Primary failure modes on granular slopes for long-legged robots are pitching and slipping \cite{mikolajczyk2022recent}. (B) Open-loop slope-adept gait and the locomotory phases. In our slope-adept gait, the CoM motion is constrained during the Single Support phase to generate no torque on the ankle joint. The CoM progression is achieved during the Double Support phase. (C) BLUEY is constrained by a planarizer, walking in the sagittal plane. The inset shows no cleat and a cleated foot. (D) 3D and untethered HECTOR robot on a granular slope testbed. }
	\label{fig:Robots}
\end{figure}

Robotic locomotion in natural environments is essential in performing operations such as exploration, deployment, and most importantly, collaboration with humans. Bipedal robots emerge as a suitable platform for these tasks, yet robust locomotion remains a central challenge due to their fragile stability: small perturbations in ground reaction forces can rapidly destabilize body dynamics \cite{atkeson2018happened,holmes2006dynamics,gu2025robust}. To mitigate this fragility, most bipedal robots rely on well-established rigid contact models and control strategies that regulate center-of-mass dynamics and ground reaction forces, assuming that terrain response is predictable and largely decoupled from body dynamics \cite{reher2021dynamic}. However, many natural terrains, such as deserts, drylands, and soil, comprise yielding substrates that deform and flow once contact forces exceed their yield stress. Granular slopes are particularly challenging for bipedal systems as foot interactions on such substrates induce excessive deformations, downward flow, and solid– and fluid–like transitions. These terrain dynamics create a strong coupling between the robot and the substrate: the robot disturbs the terrain, while the evolving terrain perturbs robot motion. In bipeds, this coupling is particularly important because their tall, upright morphology and limited foot contact area increase susceptibility to locomotor instabilities (\refig{fig:Robots} A). As a consequence, locomotor performance remains limited, and/or frequent failures occur \cite{roberts2016rhex,li2013terradynamics}. 

Challenges of bipedal locomotion on flowable slopes are partly due to the lack of efficient terrain models. Current granular terrain models, such as Resistive Force Theory (RFT), Material Point Method (MPM), and Discrete Element Method (DEM), provide useful insights for locomotion on such surfaces \cite{kamrin2024advances}. However, they involve either computationally intensive high-fidelity simulations or fast yet low-fidelity approximations. This gap limits optimization and learning-based strategies that require large-scale evaluations of robot–terrain dynamics \cite{radosavovic2024real}. Fast, low-fidelity terrain models are well-suited for such tasks, and domain randomization is widely used to reduce the sim-to-real gap \cite{li2021reinforcement,choi2023learning}. However, these methods are insufficient to capture surface deformation and flow and have not been validated on flowable slopes. Furthermore, most bipedal locomotion strategies remain ``body-centric'': they regulate robot states while treating terrain effects as external disturbances \cite{westervelt2003hybrid,ames2014human,hereid2014dynamic}. While this approach is effective on rigid surfaces, locomotion on granular slopes may benefit from a shift in perspective: regulate both robot and terrain states. Biological systems provide a successful model. Animals locomoting on deformable ground actively regulate the substrate through limb interactions for successful locomotion \cite{mazouchova2010utilization,marvi2014sidewinding}. Similarly, bipedal locomotion on flowable terrain can be enhanced by leveraging foot morphology to regulate terrain response via a ``limb-centric'' approach. 

Legged locomotion studies on granular terrain mainly focus on multi-legged robots \cite{li2013terradynamics,choi2023learning,liao2026failure}. Studies on quadrupedal robots demonstrate enhanced locomotion on sandy and muddy surfaces, using adaptive foot designs \cite{kolvenbach2022traversing,yao2024staf,shi2024foot,godon2024robotic}. Work on bipedal robots shows that informed foot morphology can facilitate locomotion in unstructured terrain \cite{piazza2024analytical,guo2020soft,tyler2023integrating}, and foot shape and intrusion depth influence the energetic performance of walking on granular media \cite{chen2025dynamic,chen2024foot}. Our prior work demonstrated stable walking with a robophysical bipedal robot on level terrain and shallow slopes \cite{xiong2017stability,gosyne2018bipedial}, and later showed that cleated feet improved walking on slopes \cite{karsai2022real}. However, a systematic characterization of biped–granular terradynamics from a limb-centric perspective and its validation in realistic, field-relevant scenarios remains lacking. 

The limitations in prior work motivate a systematic study of the terradynamics of cleated walking on flowable slopes. Here, we use a robophysical biped platform, BLUEY (Bipedal Locomotor for Understanding and Exploring Yielding terrains; \refig{fig:Robots} C). The cleats consist of rigid plates protruding perpendicularly from BLUEY's foot sole. Our experiments reveal that cleat spacing and depth are key parameters leading to diverse locomotion outcomes. We find that both sparse (12 cm) and dense (1 cm) cleat spacing can cause underperformance or failure, depending on cleat depth. In contrast, an effective cleat spacing (4 cm) enables sustained walking on granular slopes up to $30^\circ$ by distributing interaction forces to maintain substrate stresses near (or below) the yield threshold. Guided by these principles, we design robotic feet that adjust cleat depth to accommodate both rigid and flowable terrain. Further, we demonstrate that these principles extend to a large-scale dynamic bipedal robot HECTOR (Humanoid for Enhanced ConTrol and Open-source Research - \cite{li2023dynamic,kamohara2025rl} \refig{fig:Robots} D) on slopes up to $15^\circ$. Our results establish terrain-shaping contact strategies as a mechanistic foundation for robust bipedal locomotion on deformable slopes and provide a framework for integrating terradynamics into legged robot design and control. 

\section{RESULTS}
\subsection*{Cleated Foot Terradynamics Experiments}
We conducted a series of experiments to study the isolated effects of cleats on granular slopes. We used BLUEY to perform repeatable and controlled cleat-terrain interaction experiments. To this end, BLUEY's gait was designed to mitigate locomotion instabilities during cleat placement on slopes, such as pitching and slipping (\refig{fig:Robots} A). To achieve this, we regulated BLUEY’s center-of-mass trajectory using quasi-static open-loop control that minimized the robot's pitching (\refig{fig:Robots} B). First, we showed that cleats can either aid or hinder locomotion on a $20^{\circ}$ slope depending on their spacing. Second, we systematically varied cleat depth, cleat spacing, and terrain slope angle, quantified the resulting robot slippage, and identified failure modes. Third, we studied the mechanisms by which different cleat configurations facilitate or impede locomotion by conducting BLUEY experiments against a clear side wall and dual-cleat force experiments. These experiments enabled visualization of granular deformation, cleat–terrain interaction, and foot motion during slope traversal.

\begin{figure}
	\centering
	\includegraphics[width=0.99\linewidth]{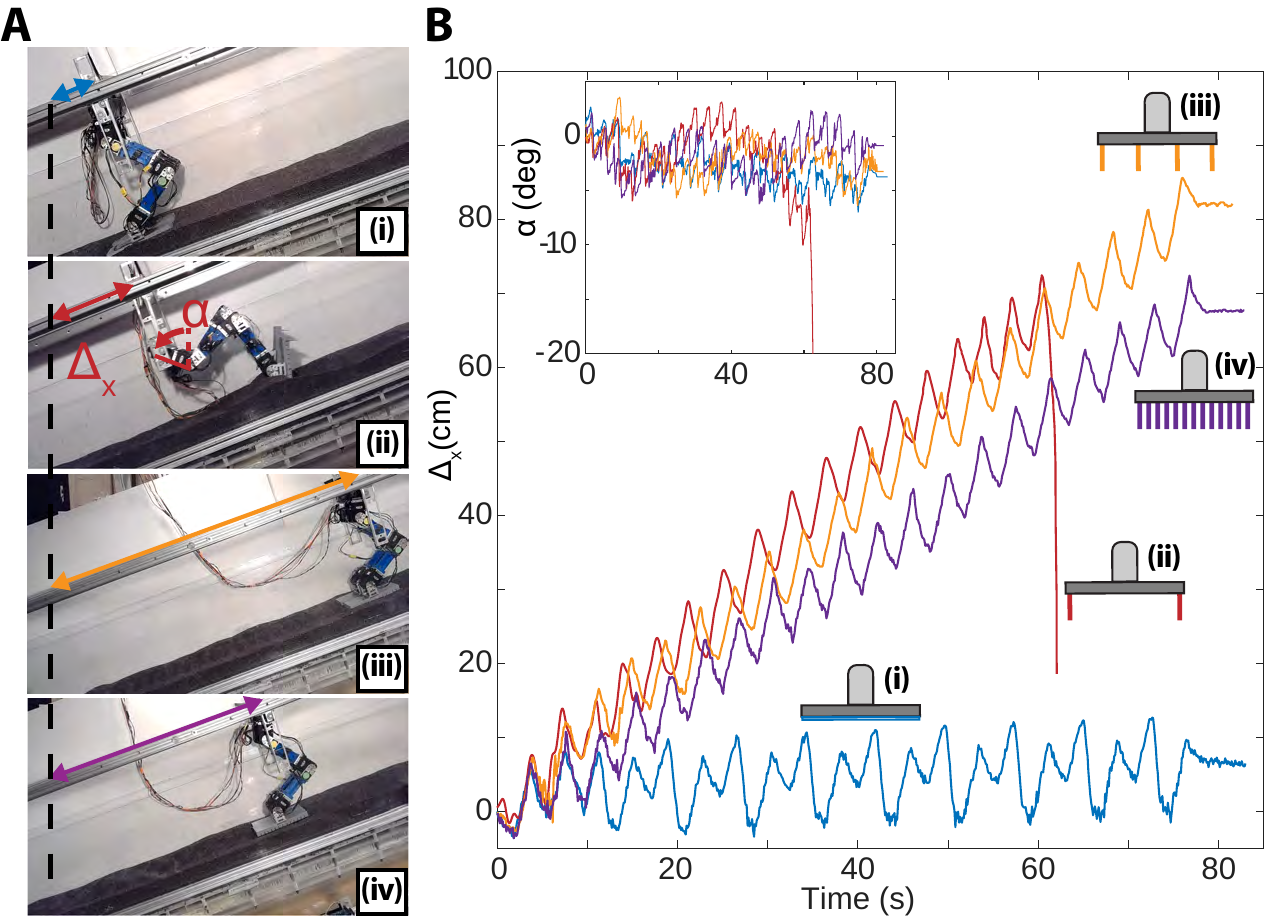}
	\caption{\textbf{BLUEY walking up a granular slope of $20^\circ$ slope with different cleated foot configurations.} (A) Snapshots of the robot during the final steps of the trial: (i) no cleat, (ii) sparse cleat (12 cm), (iii) effective cleat (4 cm), and (iv) dense cleat (1 cm) walking. The vertical dashed line represents the starting point for all the trials. (B) Displacement measurements ($\Delta_x$) relative to a commanded 100 cm walking distance are shown for each foot. The inset illustrates the pitching angle ($\alpha$) of the robot for each foot configuration. Displacement is recorded with a range sensor mounted on the testbed, and the pitch angle is recorded via an IMU mounted on BLUEY's torso. The gait parameters are kept constant across each trial: stride length of 10 cm, CoM height of 18 cm, and step period of 1 s (see Supplementary Materials for details). The testbed is prepared by introducing air through the bottom inlet between trials to reliably return the poppy seed to a loosely packed state with a volume fraction of approximately 58\%. The no-cleat case achieves 6 cm walking distance (blue curve); the sparse-cleat fails (red); and the effective (orange) and dense-cleats (purple) reach 82 cm and 67 cm walking distances, respectively. The foot configurations that achieve successful walking exhibit limited pitch angles ($\approx~6^\circ$), whereas sparse cleat spacing leads to pitching failure. }
	\label{fig:RawData}
\end{figure}

\paragraph*{Robophysical biped walking on granular slopes via cleated foot}
We measured the robot’s displacement ($\Delta_x$) and pitch angle ($\alpha$) on a 20$^\circ$ slope starting with no cleats and gradually introducing cleats with decreasing spacing  (\refig{fig:RawData}). Without cleats (blue curve), the robot slipped substantially, achieving 6$\%$ of the commanded displacement (6 cm of 100 cm). We evaluated cleated walking by first using 2 cm deep cleats mounted along both edges of the foot at 12 cm spacing (sparse cleats - red curve). The robot initially climbed up the slope, but later pitched backward and failed midway along the trackway. Next, we tested a 4 cm cleat spacing (effective cleats - orange curve) and 1 cm spacing (dense cleats - purple curve), achieving 82$\%$ and 67$\%$ of the commanded displacement, respectively. Displacement varied nonmonotonically with cleat spacing: failure occurred with sparse cleats, peak displacement was achieved at an effective spacing, and performance declined with dense cleats. All foot configurations except those with sparse cleats maintained limited pitch angles (up to ~6°), whereas sparse cleats led to excessive pitching and failure.

\begin{figure}
	\centering
	\includegraphics[width=0.52\columnwidth]{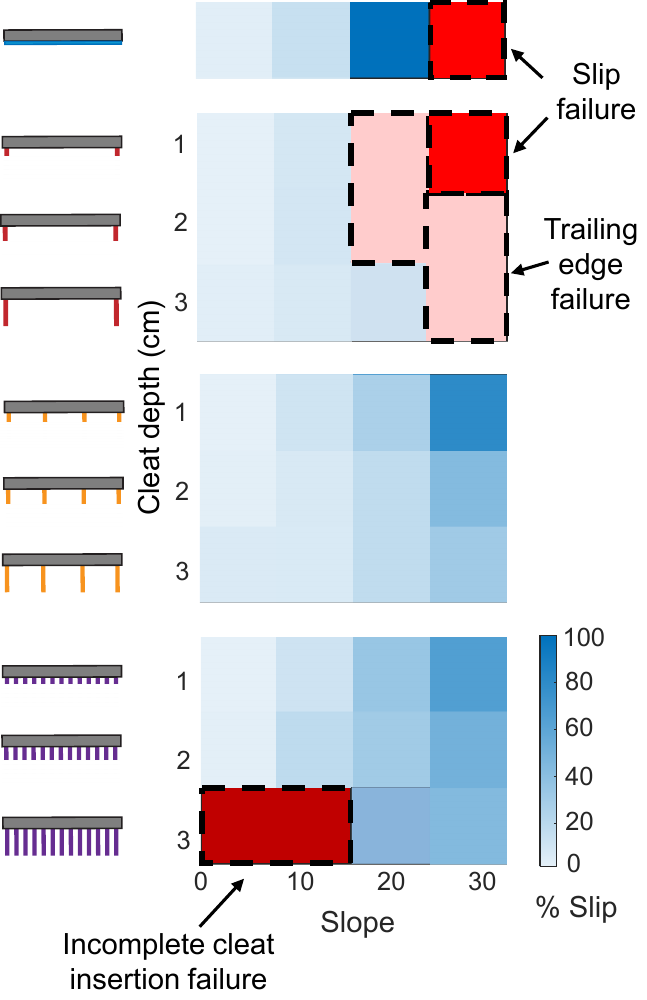}
	\caption{\textbf{Heatmap of BLUEY's slipping under varying cleat depth, spacing, and terrain slope.} The robot is programmed to walk a distance of 100 cm. The gait parameters are kept constant across trials with the same slope, and the gait for each different slope is adjusted to reduce pitching during walking (see Supplementary Material for details). Each grid component represents the mean slip value of three trials.}
	\label{fig:CleatSwp}
\end{figure}

We systematically studied the effect of cleat configurations on BLUEY's performance by varying spacing, depth, and terrain slope. A heatmap summarizing the BLUEY's slip percentage, percent of forward progress relative to commanded displacement, for no, sparse, effective, and dense cleats is shown in \refig{fig:CleatSwp}. On level and $10^\circ$ slope, almost all cleat configurations exhibited similar slip performance, ranging from $0\%$ to $10\%$, except for the dense 3 cm deep cleats, which led to failure. On $20^\circ$ slope, the sparse cleats led to locomotion failure for depths below 3 cm, and at a depth of 3 cm, the robot completed trials with $20\%$ slip. With effective cleat spacing, the robot achieved stable walking across all cleat depths, and slip improved with depth, reaching $16\%$ slip. With dense spacing, the robot achieved stable walking, but slip remained $32\%-39\%$ and did not improve with increasing cleat depth. On $30^\circ$ slope, no-cleat walking led to continuous downslope sliding and failing. Adding sparse cleats did not mitigate sliding and led to failure across all depths. At the effective cleat spacing, the robot achieved forward locomotion, and the performance improved with increasing cleat depth (up to $35\%$). With dense cleats, slip performance at 1 cm depth was better than that of effective and 1 cm deep cleats; however, deeper cleats produced small improvements in slip.
 
\begin{figure}
	\centering
	\includegraphics[width=0.99\linewidth]{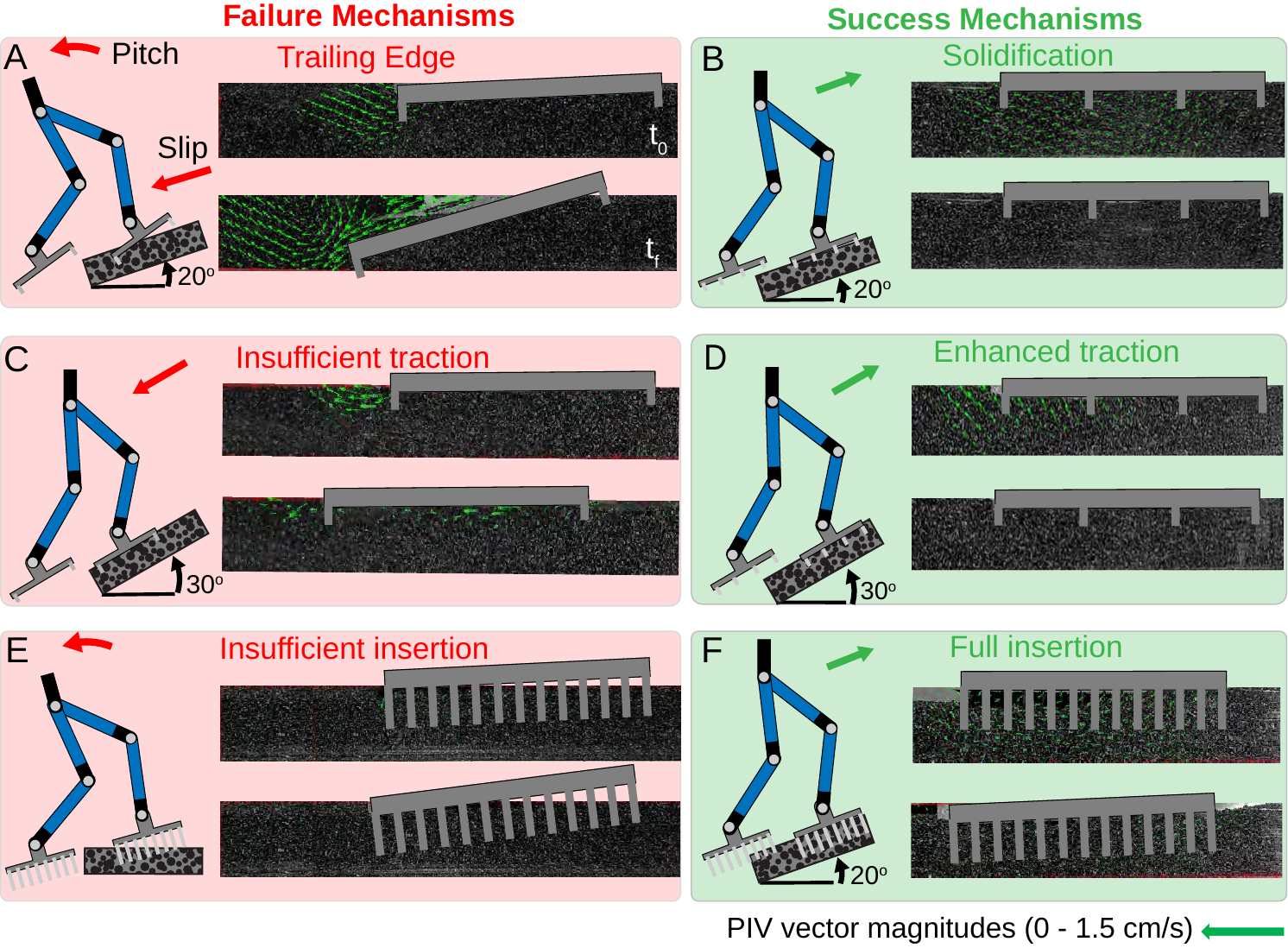}
	\caption{\textbf{Particle Image Velocimetry (PIV) images illustrating granular media - cleated foot interactions as BLUEY walks next to a clear sidewall.} The images in the left column show failure modes: (A) trailing edge, (C) insufficient traction, and (E) incomplete cleat placement. The right column shows successful walking, illustrating the flow fields associated with enhanced traction and a more solidified substrate. In (B) and (D), success is achieved with effective cleat configurations, and in (F), on steeper slopes. PIV images are captured at the peak of granular activity to highlight the flow field that leads to failure or success, with $t_0$ and $t_f$ denoting the beginning and end of particle motion.}
	\label{fig:RobotCleat_SW}
\end{figure}

We observed various failure and success cases with BLUEY depending on the cleat configuration and terrain slope. To gain visual insight into how cleats affected locomotion, we conducted experiments with BLUEY positioned against a transparent wall and visualized cleat–granular surface interactions. We used Particle Image Velocimetry (PIV) to quantify particle-flow patterns associated with representative failure and success cases from \refig{fig:CleatSwp}. PIV images in \refig{fig:RobotCleat_SW} were captured as BLUEY transitioned from double to single foot support. As the load transferred to the remaining support foot, the cleat-induced stress on the granular surface increased, generating granular activity. \refig{fig:RobotCleat_SW} A illustrates trailing edge failure on $20^\circ$, in which the robot first slipped and then pitched backward. The particle-flow field near the trailing foot edge indicates localized fluidization, characteristic of reduced granular resistance and load-bearing capacity. This reduction led to excessive backward-pitching motion. Increasing the number of cleats with effective spacing redistributed the localized particle motion uniformly beneath the foot, as illustrated by the distribution of arrows in \refig{fig:RobotCleat_SW} B. Effective cleats reduced particle motion, as no arrows were registered with the PIV images, suggesting a solid-like substrate response. At $30^\circ$ slope, sparse cleats of 1 cm depth exhibited pitching failure due to excessive slippage, as illustrated in \refig{fig:RobotCleat_SW} C. This behavior could be attributed to BLUEY's increased downward gravitational slip, exceeding the traction provided by the cleats. In contrast, effective cleats showed reduced slippage and a more uniform distribution of vectors in \refig{fig:RobotCleat_SW} D, suggesting increased traction and solid-like granular response. The PIV images in \refig{fig:RobotCleat_SW} E showed that dense cleats of 3 cm depth on level terrain did not fully penetrate the surface. In contrast, on steeper slopes, the dense cleats penetrated fully (\refig{fig:RobotCleat_SW} F). This difference could be attributed to the reduced insertion resistance on slopes.

\begin{figure}
	\centering
	\includegraphics[width=.99\linewidth]{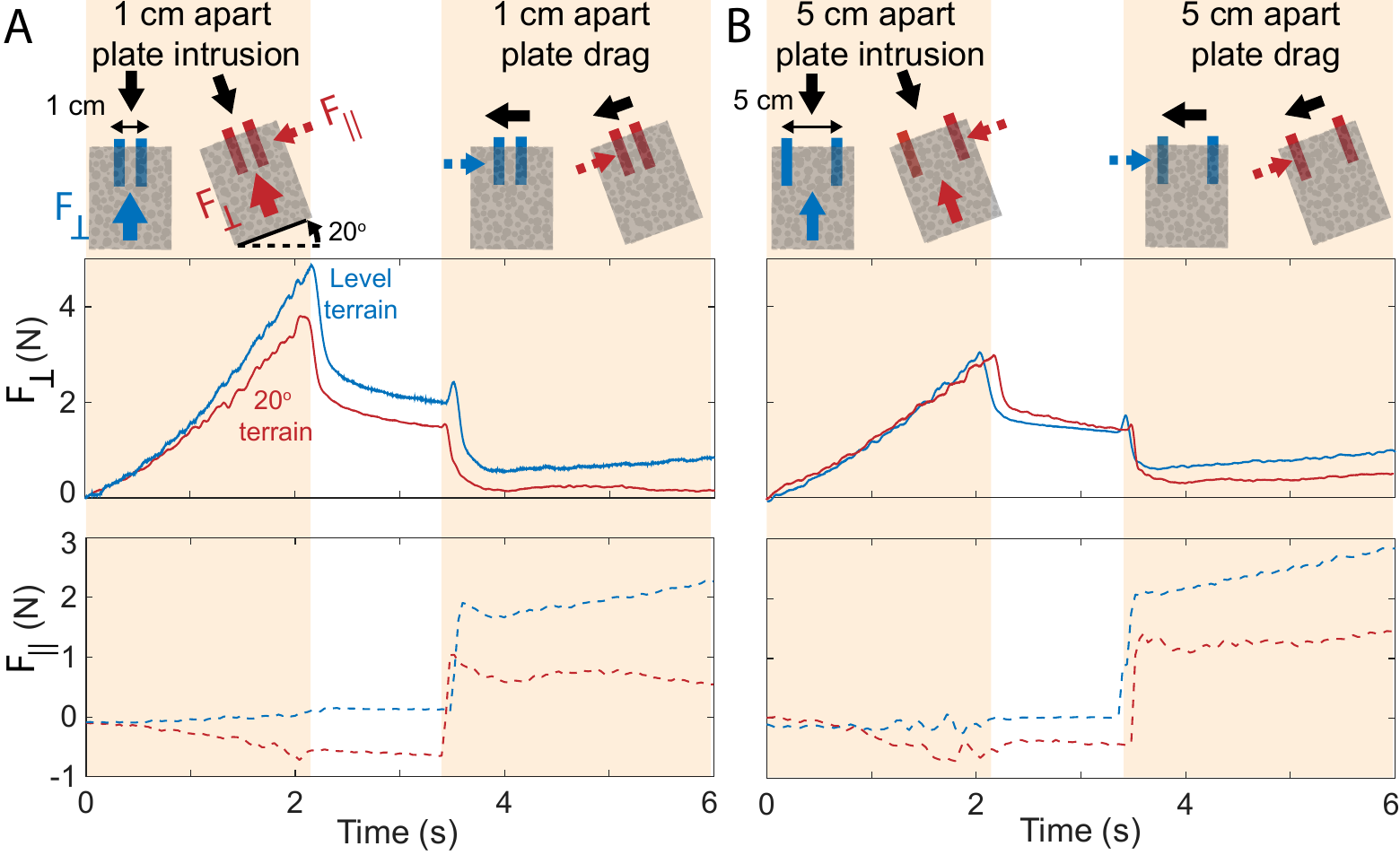}
	\caption{\textbf{Dual plate intrusion and drag experiments.} (A) Dual plates of 1 cm spacing are penetrated 3 cm deep into the granular substrate at a constant rate of 1.4 cm/s with motion direction perpendicular to the surface. After a 1-second settling period, the plates are dragged parallel to the surface in the downward slope direction. Blue and red curves represent the mean values of 3 trials for level and 20$^\circ$ incline experiments, respectively. The solid and dashed curves represent forces in the normal and tangential directions to the surface, respectively. Positive resistive forces in both directions act opposite to the direction of intrusion or drag. (B) Intrusion and drag experiments with 5 cm spaced plates. }
	\label{fig:CleatForce}
\end{figure}

\paragraph*{Granular Intrusion Experiments}
To gain insight into the granular mechanics underpinning the effect of cleat spacing on BLUEY's performance, we conducted granular intrusion tests with a pair of rectangular plates of varying spacing. The plates were first inserted perpendicular to the surface on level and $20^\circ$ inclines, representing BLUEY cleat placement. Then they were dragged down the slope, representing traction during locomotion. We used 1 cm and 5 cm plate spacings to model dense and effective cleats, respectively. \refig{fig:CleatForce} shows the measured resistive force components normal ($F_\perp$) and tangential ($F_\parallel$) to the surface. 

At 1 cm spacing, the peak $F_\perp$ of level and $20^\circ$ incline varied by only $\sim 20\%$ (\refig{fig:CleatForce} A), indicating that dense cleats on BLUEY required comparable insertion forces on both surfaces. $F_\parallel$ was approximately zero on level terrain and $\sim -0.8$ N on the incline, indicating that the sloped substrate pushed the plates downslope during intrusion. The contrasting BLUEY performance with dense and 3 cm deep cleats on level and sloped surfaces could be explained by this difference in $F_\parallel$ response. Specifically, during locomotion, this tangential, downslope pushing force likely became important when BLUEY transitioned from double to single foot support. At this moment, the single support foot, bearing the entire weight of BLUEY, caused a sudden stress increase on the surrounding terrain. Consequently, the partially penetrated cleats were pushed in a downslope direction, inducing a brief downward slip of BLUEY. This slip transiently fluidized the surrounding grains, allowing partially inserted dense cleats to penetrate fully, enabling walking.

At 5 cm spacing, the peak $F_\perp$ were approximately $ \sim 40\%$ lower than 1 cm spacing (\refig{fig:CleatForce} B), consistent with prior studies \cite{pravin2021effect,agarwal2021efficacy}. In addition, the large spacing substantially reduced cleat density: over the same span, the dense configuration would contain several more cleats than the effectively spaced configuration. Thus, effective spacing lowered both the force per plate pair and the cumulative insertion resistance of the cleated foot, making it easier to fully place the cleats. The peak $F_\parallel$ during plate insertion were comparable between the two spacings for both surfaces. However, 5 cm spaced plates exhibited higher drag resistance than the 1 cm ones, notably during steady-state drag.



\subsection*{Testing cleated-foot interaction principles on challenging conditions}
Our cleat terradynamics study revealed several key insights applicable to field-relevant terrain conditions and larger-scale bipedal robots. We first designed and built a robotic foot with retractable cleats to study locomotion and transitions between granular and rigid surfaces. Second, we implemented our cleat–terrain interaction principles on an unconstrained and autonomous robotic platform HECTOR, to facilitate walking on steep granular inclines.

\paragraph*{Robotic foot with retractable cleats}
Our studies showed that climbing steep granular slopes can be facilitated by using effectively spaced and deep cleats. However, cleats can hinder locomotion on substrates they cannot penetrate, such as rigid or obstacle-dense surfaces. To test the concept of actively regulating cleat depth depending on terrain penetrability, we developed a motor-driven cleat extension–retraction mechanism with effective spacing and integrated it into BLUEY's foot. The foot monitored the motor current as a proxy for penetrability: it retracted the cleats on rigid ground (via negative current reading) and deployed them in granular media (via positive current reading).
\begin{figure*}
	\centering
	\includegraphics[width=0.99\linewidth]{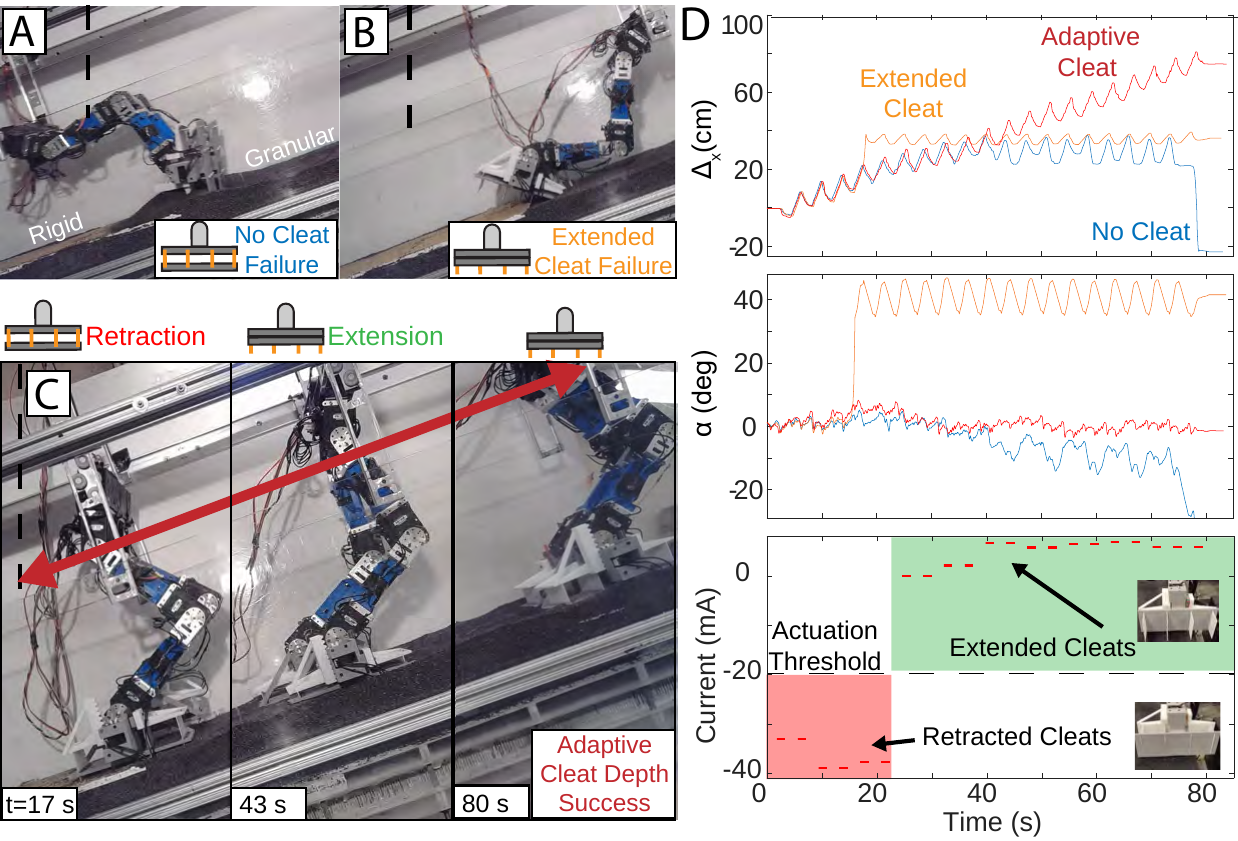}
	\caption{\textbf{Experiments with the retractable foot mechanism on rigid and granular slopes.} (A)-(B) Images of the robot failures with no and fully extended cleats, respectively. (C) Time-lapse of BLUEY walking using the adaptive cleat depth mechanism. The vertical dashed line represents the starting point for all the trials. (D) BLUEY displacement, pitch angle, and retractable cleat-foot actuator current reading. The actuated cleats facilitate locomotion on complex surfaces: granular and rigid terrains.}
	\label{fig:ActCleats}
\end{figure*}
We evaluated three foot configurations on a $20^\circ$ slope: no cleats, extended cleats, and adaptive cleat-depth control. Starting from a rigid surface, we defined successful locomotion as the ability to transition and sustain walking on the granular surface. The resulting performance measurements are presented in \refig{fig:ActCleats}, illustrating the displacement and body pitch angles. While the no-cleat BLUEY successfully transitioned from the rigid to the granular surface, it slipped and pitched backward on the granular surface (\refig{fig:ActCleats} A). With the cleats fully extended, the robot failed to transition onto the granular surface, pitching during the attempt (\refig{fig:ActCleats} B). For the actuated-cleat trial, the cleats were initially retracted, leaving only a 3 mm tip extension to probe the ground during contact. While walking on the rigid surface, the cleat tip contact generated a negative current on the cleat extension motor (measured only during the stance phase of walking). The controller then used this signal to prevent cleat extension. Upon transitioning to the granular surface, the sensed current approached zero, triggering cleat extension of 2.5 cm. In this mode, the robot transitioned and completed the remaining granular segment, with minimal variation in the pitch angle (\refig{fig:ActCleats} C).

\begin{figure*}
	\centering
	\includegraphics[width=0.99\linewidth]{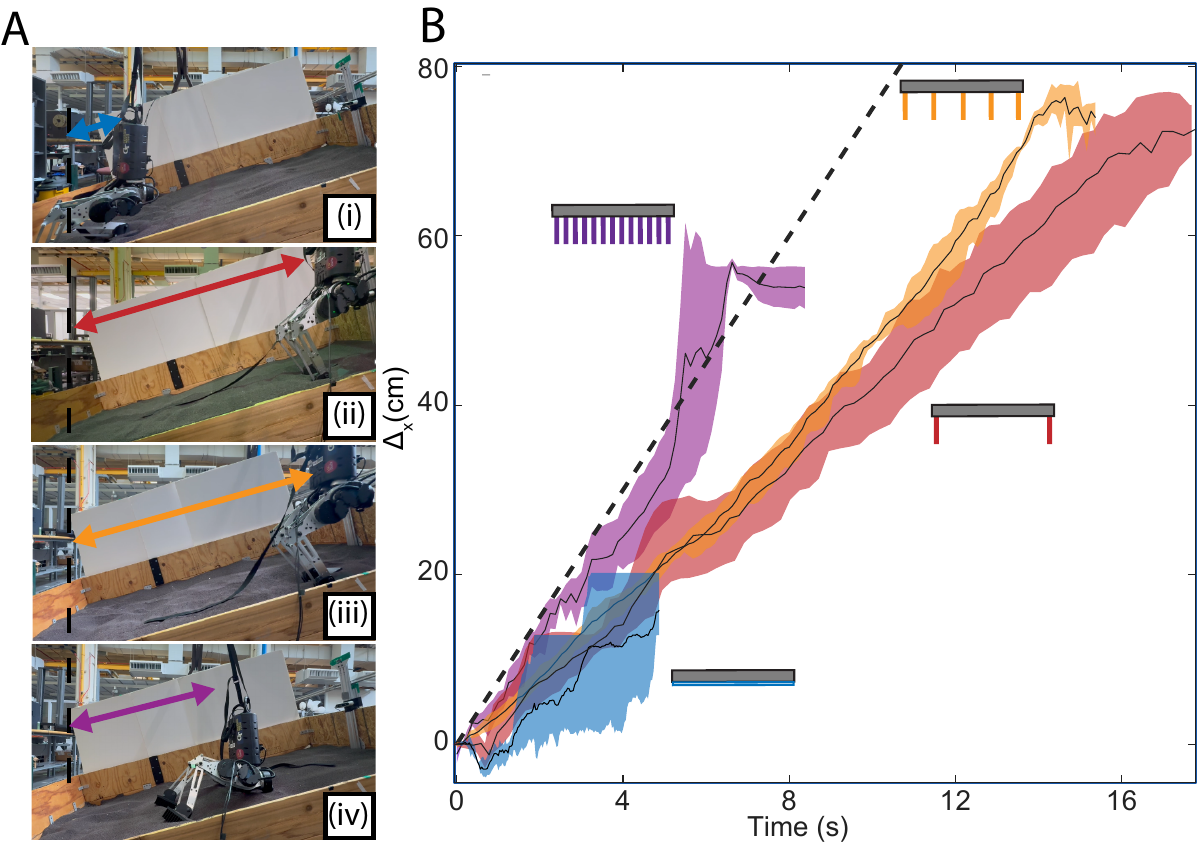}
	\caption{\textbf{HECTOR locomotion experiments with different foot configurations.} (A) Snapshots of HECTOR robot walking with (i) no cleat, (ii) sparse cleat,  (iii) effective cleat, and (iv) dense cleat configurations. HECTOR is initiated to walk at the positions depicted by dashed vertical lines in all trials. (B) The displacement performance of HECTOR under closed-loop MPC control. The displacement is captured by a depth camera attached to the distal end of the testbed with a resolution of $\pm 0.5$ cm. The maximum displacement variations of the shaded regions are 20 cm, 12 cm, 4 cm, and 28 cm, for no, sparse, effective, and dense cleats, respectively.}
	\label{fig:Hector_fig}
\end{figure*}

\paragraph*{Using cleat terradynamics insights for 3D untethered bipedal walking} 
We tested the scalability of cleat–foot and terrain interactions using HECTOR, an unconstrained 3D biped that is twice as tall and $10$ times heavier than BLUEY. We conducted experiments on a $15^\circ$ granular incline using foot configurations similar to BLUEY: no cleats, sparse cleats (16 cm spacing), effective cleats (4 cm spacing), and dense cleats (1 cm spacing). We modified the HECTOR foot by creating a planar sole to accommodate cleat designs similar to BLUEY. We increased the foot width in the frontal plane to improve lateral support. This modification increased the foot's contact area in the lateral direction for enhanced lateral stability during locomotion. We set the cleat depth to 4 cm, the depth that could be accommodated without causing excessive ground interference during foot swinging. 

HECTOR was operated using a single-rigid-body-dynamics-based model predictive controller \cite{li2023dynamic}. We commanded the robot to follow a predefined speed of 7.5 cm/s for all foot configurations. \refig{fig:Hector_fig} illustrates snapshots of successful and failed walking, along with the achieved distances for these cleat configurations. Feet without cleats failed after only a few steps (blue curve). With sparse cleats (red curve), the robot achieved stable dynamic locomotion; however, performance varied widely across trials, and the robot underperformed in forward speed and terrain traction. Dense cleats (purple curve) enabled the robot to track the commanded velocity initially, but later led to failure. Effectively spaced cleats (orange curve) yielded the best overall performance: although the robot still exhibited moderate velocity-tracking errors, the cleated foot reduced velocity fluctuations and improved consistency.

The HECTOR experiments showed that cleat–granular interaction principles could be scaled to larger and heavier systems. Through our terradynamics studies, we distilled key cleated foot design insights that facilitate robust locomotion in challenging real-world conditions. First, cleats should be spaced to minimize cooperative interaction effects during intrusion. The effective spacing reduces agitation on flowable slopes during cleat placement, solidifying the flowable terrain beneath the foot and decreasing flowability. Second, cleats should be sufficiently deep to minimize slippage during locomotion phase transitions. As our cleat depth intrusion experiments (\refig{fig:CleatDepth}) and prior work (\cite{marvi2014sidewinding}) showed, the drag forces on penetrated plates increase approximately linearly with depth. Therefore, deeper cleats generate higher drag resistance, reducing the risk of excessive slip on flowable slopes.

\section{DISCUSSION}
We studied how cleated-foot interactions on granular slopes affect bipedal locomotion by systematically varying cleat spacing, depth, and surface slope of a robophysical robot, BLUEY. We first measured BLUEY's displacement and pitch angle and observed particle velocities beneath the cleated foot. We then quantified the granular resistive forces of varying cleat spacings by approximating the cleated foot as dual plates. Finally, we validated our findings in challenging scenarios by (i) developing a robotic foot capable of varying cleat depth that adapts to rigid and granular terrain and (ii) scaling the derived principles to an unconstrained bipedal robot, HECTOR.

\paragraph{Cleated foot -- soft matter interactions}
Our studies revealed a set of cleated-foot interactions worth discussion and further investigation. The BLUEY experiments showed that bipedal walking on granular slopes is highly sensitive to cleat density and depth. We observed locomotion stagnation without cleats on a 20$^\circ$ slope. We hypothesize this could be due to the foot–surface friction being below or equal to the downslope gravitational pull. 

\textbf{Sparse Cleats:} The inclusion of sparse cleats with a depth of 2 cm failed, despite our expectation that an increase in foot-surface friction would enhance locomotion. PIV images showed that this failure was triggered by the downward drag motion of the trailing cleat. Our dual plate experiments and prior studies \cite{gravish2014force} indicate that dragged plates lead to material fluidization near them. Here, the trailing cleat acted as a dragged plate during BLUEY's downward slip, locally weakening the substrate at the trailing edge. As a result, the fluidized trailing region could not sustain external load, causing BLUEY to pitch backward. Increasing cleat depth reduced the initial slip and mitigated this failure mode. 

\textbf{Dense Cleats:} With deep and dense cleats, BLUEY exhibited slope-dependent performance: cleats led to failure on level/shallow slopes and enabled walking on steep slopes. PIV analysis showed that deep and dense cleats did not fully insert into the level terrain during walking. This could be due to increased insertion forces arising from narrow cleat spacing and a higher cleat count (1 cm spacing feet had 12 cleats). Scaling peak insertion forces of two narrow plates to 12 cleats suggests that ~30 N was required for full penetration of the densely cleated foot, exceeding BLUEY’s weight (14 N). As a result, BLUEY walked on partially penetrated cleats, which induced pivoting of the CoM, eventually leading to pitching backward and failure. However, BLUEY achieved walking on steep slopes by fully penetrating the same cleat configurations, despite a scaled force of ~24 N required to fully insert them. This behavior likely arose because, on sloped terrain, BLUEY experienced downward slip motion due to gravitational pull that dragged the partially penetrated dense cleats down, whereas on level terrain, the cleats were not dragged. This slope-induced drag led to local yielding of the surface around the cleats, and a transient fluidization, enabling full cleat penetration on slopes, consistent with the particle flow patterns observed in \refig{fig:RobotCleat_SW} F. 

\textbf{Effective Cleats:} We found that cleats enhanced BLUEY's locomotion most when they were effectively spaced and penetrated sufficiently deeply. Effective spacing minimizes overlap of penetration zones, reducing inter-cleat coupling forces during insertion. Reduced insertion resistance enables cleat placement with less disturbance to the terrain, keeping flowable slopes below or near the yield stress under load. This could result in less downward granular flow as BLUEY walks up hills. This observation is consistent with prior studies, which show that successful robotic ascent on yielding surfaces is achieved by minimizing substrate disturbance \cite{marvi2014sidewinding, mazouchova2013flipper}. Additionally, PIV images suggest that effectively spaced cleats recruit and solidify the entire granular volume beneath the foot during walking, rather than mobilizing particles only around individual cleats. Hence, successful locomotion on slopes may not only depend on generating traction forces via cleats but also on creating a uniformly solidified region beneath the foot. BLUEY experiments supported this observation: sparse, 3 cm deep cleats generated large traction on 30° slopes yet led to failure, whereas effective and 1 cm spaced cleats generated small traction yet enabled successful locomotion. These behaviors suggest an analogy to aerodynamic design, where appropriately shaped airfoils achieve stable, low-energy flight by organizing flow and minimizing detrimental interactions such as separation and turbulence \cite{he2023aerodynamics}. Similarly, effective cleat spacing appears to ``condition'' the granular substrate. In this sense, cleat geometry and spacing act as a form of terradynamics shaping, enabling robust locomotion not by increasing traction force alone, but by structuring the interaction with the medium. 

\paragraph{Scaling cleated-feet interaction principles}
Although HECTOR and BLUEY differ substantially in mass, kinematics, and gait, both systems exhibited similar sensitivity to cleat spacing on steep granular slopes (15$^\circ$--20$^\circ$). For both robots, effective cleat spacing led to improved performance compared with sparse and dense cleats. The difference between the two robots emerged in the no-cleat feet. BLUEY sustained locomotion despite substantial slip on $20^\circ$, whereas HECTOR exhibited rapid failure on $15^\circ$. We attribute this difference to the gait dynamics and robots' kinematics. HECTOR’s dynamic and unconstrained walking struggled to make forward progress with low-traction feet, and its heavy weight more readily fluidized the substrate. In contrast, BLUEY’s quasi-static gait, lower weight, and planarized structure maintained foothold stability even under large slippage.

We note that effective cleats facilitate more effective use of sensory feedback in HECTOR’s MPC controller by creating solid-like terrain response. Specifically, granular terrain can transition to a fluid-like state under complex forcing, reducing load-bearing capacity. Existing sensing modalities and terramechanics models remain limited in predicting granular terrain yielding and deformation in real time, which can cause large fluctuations in sensor readings and lead to failure \cite {roberts2016rhex,kamrin2024advances}. As a result, the MPC's corrective actions may be insufficiently matched to the evolving terrain state, leading to the rapid failures as observed in the no-cleat walking. We conjecture that effective cleats limit terrain yielding by (i) trapping particles uniformly into load-bearing regions beneath the foot (ii) distributing the granular flow across the entire foot sole. These mechanisms limit fluidization beneath the foot and facilitate a more solid-like granular response. Consequently, this leads to enhanced load-bearing capacity and allows HECTOR’s sensory feedback mechanisms to operate under more nominal conditions.

Our study revealed several key interaction principles governing locomotion on yielding slopes. These principles may provide a practical framework for extending cleat-assisted locomotion to larger bipedal systems. Rather than relying on a specific robot morphology, our results suggest that locomotion on granular slopes can be achieved by selecting cleat morphology that uniformly distributes interaction forces while avoiding excessive substrate yielding. Although our experiments were conducted on a small-scale biped, the underlying mechanism can be scaled for larger systems. Specifically, HECTOR cleats were scaled to have approximately three times the surface area of the deepest effective BLUEY cleats, facilitating locomotion in a robot that was roughly ten times heavier and twice as tall as BLUEY. For even larger bipeds, increased body weight will likely require larger feet, deeper cleats, or alternative cleat arrangements to preserve these interactions. Future studies combining large-scale experiments, terradynamic modeling, and DEM simulations could establish scaling laws linking robot size, substrate properties, and cleat geometry, enabling the design of effective feet for human-scale robots operating on deformable terrain.

\paragraph{Study limitations}
We acknowledge several limitations of our study. We considered only rectangular cleats with a maximum depth of 3 cm, as constrained by the dimensions of BLUEY. Similarly, for HECTOR, we only used 4 cm deep cleats. We also limited cleat intrusions to an angle normal to the surface and varied cleat spacing along only one axis. Future work could study alternative cleat geometries, varying intrusion angle, spacing, and depth to determine whether these design choices create a more solid-like terrain response. Moreover, we used a modified ZMP-based gait on BLUEY, operated within a narrow parameter space, designed to place cleats to minimize terrain disturbance. Future work could explore optimized gait strategies (e.g., step timing, step length, foot attack angle). It is worth noting that, despite the improvements provided by effective cleats, there remains a limit to the achievable performance through spacing and geometry alone. Because steep, flowable slopes undergo localized yielding and a fluid-like state under contact, foot–terrain contact cannot remain perfectly rigid during walking. Consequently, even with optimized cleats, the robot will typically exhibit nonzero slip. 

While wall-PIV and dual-plate granular force measurements provide valuable insights, they exhibit global granular behavior and cannot quantify particle-scale mechanisms. Future Discrete Element Method (DEM) study of multiple intruders on slopes could quantify jamming/force-chain structures and other local state variables as a function of intruder kinematics and spacing. This could clarify why reduced spacing amplifies granular intrusion forces by visualizing and quantifying the coupled interactions among multiple intruders. Furthermore, our study was limited to 1-mm-diameter poppy seeds. Particle size, shape, and density, such as fine or coarse sands, may alter the optimal cleat spacing. Future studies combining DEM simulations with systematic experimental validation across a range of granular substrates will help determine how particle properties influence effective cleat design.

Our demonstrations with the actuated cleats considered transitions primarily from large, rigid surfaces. Future work will investigate the robot’s ability to transition from small, partially embedded, or visually occluded obstacles, reflecting natural conditions. In addition, HECTOR’s MPC controller has limitations in the present study: we evaluated performance over a limited traversal distance and explored only a narrow subset of controller and gait parameters. More extensive field testing on slopes larger than 15$^\circ$, potentially coupled with learning-based approaches (e.g., reinforcement learning \cite{radosavovic2024real}), may yield more robust and higher-performing gait strategies for autonomous locomotion. 




\section*{MATERIALS AND METHODS}
\subsection*{Robot Kinematics} 
 \textbf{BLUEY}: BLUEY is constructed using 6 Dynamixel MX-64 Servo motors, composed of ankle, knee, and hip joints connected with 3D printed links. The robot consists of 15 cm leg links and a 10 cm torso. Each foot is mounted 4 cm below the ankle joint. The robot's total height is $44$~cm, and the weight is $1.4$~kg. The width of the robot is 12 cm. The robot is constrained to the sagittal plane using rollers attached to its torso, which roll along vertical rails mounted on horizontal rails. This enables the robot to translate and rotate freely in the sagittal plane.

\noindent
\textbf{HECTOR}: The autonomous biped has a standing height of 85 cm, weighs 15 kg, and consists of legs and a torso. Each leg consists of 5 degrees of freedom with hip yaw, hip roll, thigh, knee, and ankle joints, driven by quasi-direct-drive actuators with a single-stage 9.1:1 planetary gearbox. Details can be found in \cite{li2023dynamic}.

\noindent
\textbf{Feet and Cleats}: Each BLUEY foot is 15 cm long and 5 cm wide and contains 13 slits; the cleats are inserted into these slits beneath the foot. The BLUEY cleats are rectangular carbon-fiber inserts (5 cm wide, 1 mm thick) with protrusion depths ranging from 1–3 cm. BLUEY cleat spacing was varied between 1 cm, 4 cm, and 12 cm, and cleat depth was varied from 1 cm to 3 cm. HECTOR uses cleats that are 3D-printed as an integral part of the foot and incorporated into the robot’s ankle–foot assembly. Each HECTOR cleat is 8 cm wide, 4 cm deep, and 1 mm thick, and the HECTOR foot measures 17 cm × 10 cm (length × width). HECTOR cleat spacing was varied between 1 cm, 4 cm, and 16 cm.

\subsection{Bipedal Gaits}
\textbf{BLUEY}: The locomotory phases of bipedal systems consist of ``Single Support (SS)'' and ``Double Support (DS)'' phases. In the SS phase, the robot swings one leg while the other remains in contact with the ground. In the DS phase, both legs are in contact with the ground. Bipedal robot stability becomes particularly critical during the SS phase on granular slopes due to the limited number of ground contacts. Even small terrain disturbances during locomotion can lead to failure as the substrate yields and flows downward. Specifically, pitching instability is a common failure mode in high–center-of-mass (CoM) systems, and bipeds are inherently limited in their ability to lower their CoM to mitigate pitching failure. We developed an open-loop gait for BLUEY based on the ZMP framework, adapted for sloped terrain to reduce robot pitching, following approaches in \cite{kajita2003biped,xiong2017stability}. Our slope-adept gait first constrains the moment point during the SS phase—and therefore the center-of-mass (CoM) trajectory—to generate zero ground reaction moment at the support ankle (Fig. 1 B). We then rotate the CoM position by an angle equal to the terrain slope in the opposite direction so that it remains aligned with the gravity vector regardless of the slope. We define $p_{CoM}^0 = [y_{CoM}^0,~z_{CoM}^0]^T$ as the CoM position of the robot on level terrain at the onset of the SS phase. For a terrain slope of $\theta$ in counterclockwise direction, we rotate $p_{CoM}^0$ in the clockwise orientation with
\begin{equation}
 p_{CoM}^{\theta} = R_x(\theta)p_{CoM}^0
\end{equation}
where $R_x(\theta)p_{CoM}^0\in \mathbb{R}^{3\times3}$ is an orientation matrix. Constraining the ZMP region to the centroid of the support foot during SS gives:
\begin{equation}
 P_{ZMP}^{SS} = \frac{\tau}{mg}=0,
\end{equation}
where $m$ is the robot's weight, $g$ is the gravitational force, and $\tau$ is the ground reaction torques on the ankle. Thus, regardless of the robot's configuration and the terrain slope, the robot acts as an upright stable inverted pendulum during the SS phase (see \refig{fig:Robots}). This helps minimize pitch caused by unpredictable terrain deformation during SS and enables more parallel foot contact on granular slopes.

After taking a step, we then translated the CoM position forward during the DS phase by an amount of half step length $L$, with the corresponding ZMP given as
\begin{equation}
 P_{ZMP}^{DS} = L/2,
\end{equation}
The DS phase facilitates net forward displacement, with the robot having a larger support polygon due to its forward and hind foot contacts. The gait scheme continuously alternates between “static CoM in SS” and “progressive CoM in DS” phases. 

To reduce inertial effects during slope locomotion, we operated in a quasi-static regime by setting the step period to 4 s. To prevent foot and cleat scuffing, we used sufficient cleated-foot clearance during the SS phase and inserted/extracted the cleats normal to the slope surface, minimizing granular reorganization during cleat placement and extraction. The gaits were generated with a step length of 10 cm for 10 steps, assuming that the robot was sunk 1 cm into the granular media. These gaits were generated by manually tuning the trajectories for the terrain slopes tested in our experiments (Figs. 2–4). The torso angle was maintained at 0° relative to the vertical axis across all slope conditions.

The gait trajectories were generated using MATLAB, and individual controllers on each servo handled position control of the associated joint. The inverse kinematic solver used the CoM trajectory to generate the angular profiles of each joint. These joints were controlled using the PID controllers built into the motor drivers. 

\textbf{HECTOR}: The unconstrained biped HECTOR was modeled using single-rigid-body dynamics (SRBD) and controlled with a model predictive controller (MPC) that maps control inputs to joint torques via the contact Jacobian. The nominal gait uses a 0.2 s step period; to reduce dynamic effects on slopes and leverage insights from our BLUEY gait, we increased the step period to 0.55 s.

\subsection*{Experimental Procedure} 
\textbf{BLUEY Experiments}:
Robophysical locomotion experiments with BLUEY were performed using a tiltable air-fluidized bed of poppy seeds (1 mm diameter) with dimensions of 120 cm × 25 cm and a granular depth of 12 cm (\refig{fig:SI_BlueyTestBed}). Poppy seeds serve as a model substrate for flowable and deformable terrain, such as sand and soil. Before each experiment, the bed was air-fluidized to reset the poppy seeds to an initial loosely packed state with a volume fraction of 58\%. Air-fluidization enabled a uniform, loosely packed test surface (see Fig. \refig{fig:Robots}). The terrain angle was varied from 0° to 30° in 10° increments. The weight of the rails used to planarize BLUEY was offset by attaching appropriate counterweights to the rails at each terrain slope. The testbed incorporates a range sensor, mounted against the far end of the trackway to measure the distance of the robot's torso during walking. We captured the robot's pitch angle by mounting an IMU on the torso. The distance between the robot limbs and the bed boundary was maintained at 5 cm, sufficient to avoid boundary effects \cite{mazouchova2013flipper}. Each experiment began by setting BLUEY to the prescribed initial posture for the target slope angle, followed by placement on the testbed. All gaits were designed such that the torso remained upright, with a 0$^\circ$ orientation relative to gravity. The IMU readings were then checked to confirm that the torso angle was close to 0°. Walking was subsequently initiated, and both the robot displacement and torso pitch during locomotion were recorded.

\textbf{HECTOR Experiments}:
Locomotion experiments were performed using a tilted testbed of 15$^\circ$, consisting of poppy seeds with an approximate volume fraction of 61\% \refig{fig:SI_HectorTestBed}. The bed was 180 cm long, 80 cm wide, and filled with poppy seeds to a depth of 30 cm. Before each experiment, we raked the poppy seed test bed. Hector was first set in standing mode and consequently commanded to walk up slopes. We commanded the robot to walk at a constant velocity of 7.5 cm/s and manually halted locomotion when the robot reached the end of the testbed. We used an Intel RealSense depth camera mounted at the end of the testbed to register the displacement of the robot. 

\subsection{Granular Intrusion Apparatus}
Intrusion experiments were conducted in an air-fluidized testbed measuring 60 cm in length, 30 cm in width, and 20 cm in depth, filled with poppy seeds (\refig{fig:SI_DualPlates}). Two Firgelli linear actuators, each mounted on a separate linear stage perpendicular to each other, were used to intrude and drag the cleats at a speed of 1.4 cm/s. The cleats were machined from aluminum and measured 3 cm in depth, 5 cm in width, and 1 mm in thickness. Each cleat was attached to an aluminum holder connected to an ATI Mini40 SI-80 force/torque sensor. The actuators were controlled using LabVIEW, and force data were acquired at 1 kHz through an NI DAQ system.

\subsection{Actuation of Cleated Foot}
The robotic foot system that actuates cleats comprises two 3D printed components: the base component and the cleat component \refig{fig:SI_ActCleats}. The base serves as a structural frame and is attached to the robot's ankle joint. It functions as a flat foot with slits on its bottom to allow the movement of the cleat component. The base component houses a rotary motor, Dynamixel XC330, which is fixed to the structure. The shaft of this motor houses a pinion. The cleat component consists of vertical plates with an effective spacing, each measuring 3 cm in depth, 5 cm in width, and 1 mm in thickness. This component incorporates a vertical rack that interacts with the motor pinion to facilitate linear motion. When the motor is actuated, the pinion’s rotation translates into linear vertical motion through the rack of the cleat component, allowing the cleats to extend or retract. Both components are connected via a linear bearing to stabilize linear cleat movement.

To sense ground contact, the cleats were initially commanded to extend to a depth of 3 mm during the single support phase. Contact sensing was performed only during the double-support phase and only with the newly contacted foot. The contact with a rigid surface increases the load on the motor due to the extension of the cleats. The motor’s internal PID controller resists this motion, leading to increased current readings (negative values). In contrast, on soft terrain, such as poppy seeds, the cleats penetrate the material with minimal resistance, resulting in current readings that remain near zero or positive. By applying a simple threshold to the motor current, we distinguished between penetrable and non-penetrable surfaces. The cleats were fully extended on penetrable surfaces, whereas no action was taken on rigid surfaces. We used a sandpaper-covered surface to create a high-friction and rigid slope. A pseudocode is given below for the sensing and decision-making.

\begin{algorithm}[t]
\caption{Ground Contact Sensing and Cleat Depth Control}
\begin{algorithmic}[1]
\REQUIRE $\ell_{\text{probe}} = 3~\text{mm}$, $\ell_{\max}=2.5 cm$, $\varepsilon$
\STATE Initialize $\ell \leftarrow \ell_{\text{probe}}$
\FOR{each locomotion cycle}
    \IF{phase == SINGLE\_SUPPORT}
        \STATE $\ell \leftarrow \ell_{\text{probe}}$ \COMMENT{Set probe length}
        \STATE set cleat length $\ell$
    \ENDIF
    
    \IF{phase == DOUBLE\_SUPPORT}
        \IF{foot == leading\_foot}
            \STATE $I \leftarrow$ read foot actuator current
            \IF{$I \geq -\varepsilon$}
                \STATE $\ell \leftarrow \ell_{\max}$ \COMMENT{Penetrable terrain}
                \STATE set cleat length $\ell$
            \ELSE
                \STATE $\ell \leftarrow \ell_{\text{probe}}$ \COMMENT{Rigid terrain}
            \ENDIF
        \ENDIF
    \ENDIF
    
    \STATE execute locomotion step
\ENDFOR
\end{algorithmic}
\end{algorithm}

\section*{Supplementary Information}
\section*{Supplementary Text}
\subsection*{Effect of cleat depth on granular forces}
\begin{figure}
	\centering
	\includegraphics[width=0.99\linewidth]{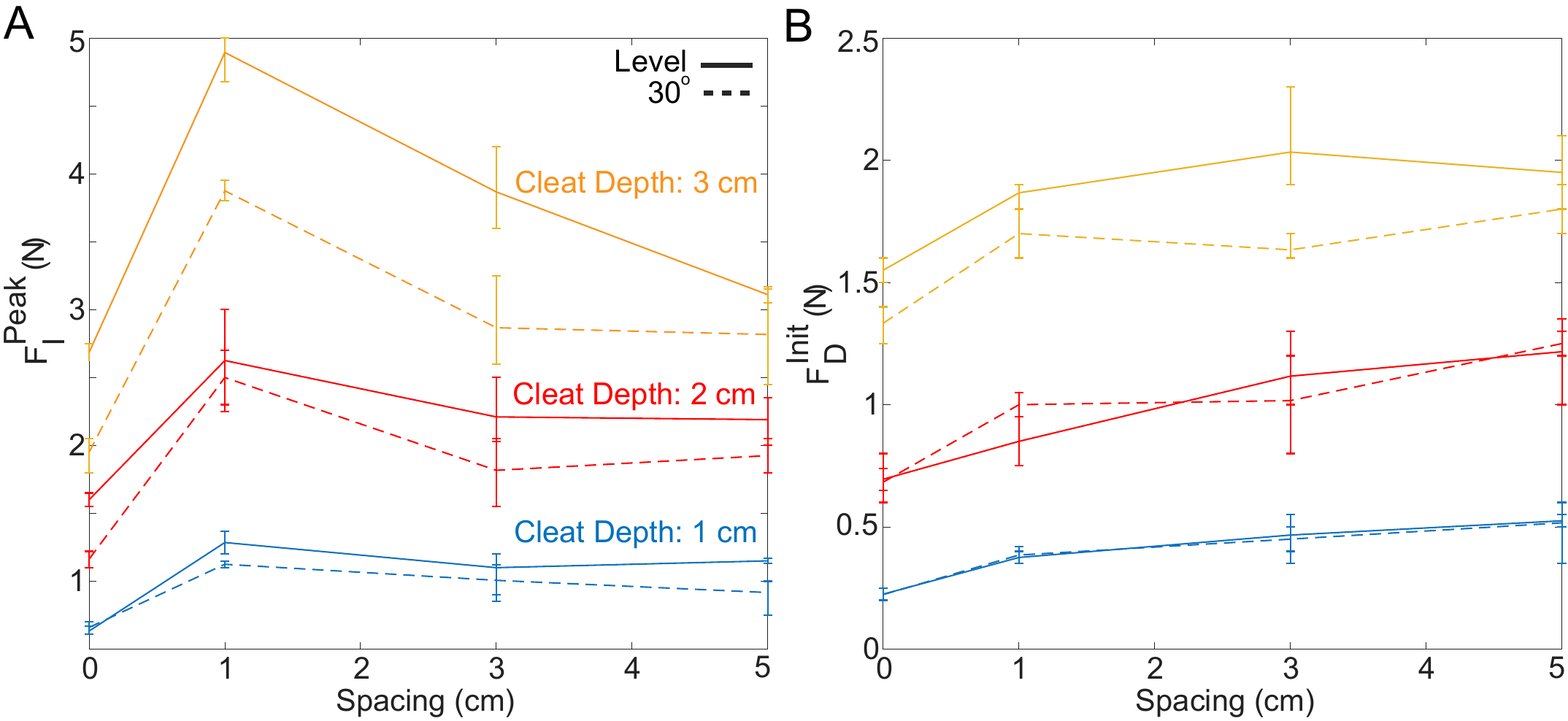}
	\caption{\textbf{Intrusion and drag of dual plates of varying spacing and depth.} A) illustrates the peak granular resistive forces that the dual plates experience during intrusion on level and 30$^\circ$ inclines. B) illustrates the initial peak resistive forces experienced by the dual plates as they are dragged down the slope.}
	\label{fig:CleatDepth}
\end{figure}

The effect of cleat depth on BLUEY’s performance varies with cleat spacing. For sparse and effective spacing, deeper cleats improved locomotion performance. In contrast, for dense spacing, deeper cleats led to failure on shallow slopes while enhancing performance on 20$^\circ$ and 30$^\circ$ inclines. Prior studies have demonstrated that granular forces strongly depend on the intruder's geometry and depth \cite{askari2016intrusion,li2023dynamic}. To quantify the mechanisms underlying these observations on slopes, we conducted dual-plate intrusion experiments, measuring resistive forces as a function of spacing and insertion depth on both level ground and a $30^\circ$ incline. We first inserted the plates perpendicular to the surface and measured the peak intrusion force, $F_I^{\text{Peak}}$, to evaluate how intruders cooperatively disturb the material (Fig.~\ref{fig:CleatDepth} A). Increased resistive forces could induce more particle mobilization and downward flow. We then dragged the submerged plates down the slope and quantified the drag resistance of the material before yielding, i.e., initial drag $F_D^{\text{Init}}$, to assess how plates affect granular traction (Fig.~\ref{fig:CleatDepth} B). Higher initial drag resistance could correlate with BLUEY's improved resistance to slipping downward. 

As the spacing between the plates increases, the intrusion forces exhibit nonlinear behavior, with zero spacing corresponding to plates touching and 5 cm to the largest separation. $F_I^{Peak}$ starts with the lowest forces at zero spacing, reaches a maximum at 1 cm separation, and settles between 4-5 cm separation to a value that is slightly larger than zero spacing forces. This trend persists on $30^\circ$ slopes and across all cleat depths, with reduced force magnitudes. However, we observed that as depth decreased, the cooperative forces became less sensitive to increasing separation. These observations are consistent with our robophysical experiments, which showed that dense and deep cleats were increasingly difficult to penetrate. As illustrated in Fig. 3, BLUEY traversed all tested slopes with dense cleats up to 2 cm in depth, penetrating them during the locomotion cycle. Scaling 1 cm spaced dual-plate force measurements to the BLUEY foot at a depth of 2 cm yields an estimated insertion force of ~15 N, which closely matches the force BLUEY can generate to successfully place its cleats. However, beyond this depth, BLUEY was unable to achieve full cleat penetration, resulting in failure.

The initial drag resistance exhibits a near-monotonic increase with plate spacing for both level and $30^\circ$ incline. Additionally, increasing cleat depth increases $F_D^{\text{Init}}$, consistent with prior observations of depth-dependent drag in granular media \cite{gravish2014force}. These trends show competing effects from a locomotion perspective, necessitating a trade-off between cleat spacing and depth. The increasing cleat spacing enhances shear resistance during drag, which can reduce BLUEY's slip, but it also decreases the total number of cleats on the foot, reducing overall ground contact. Conversely, decreasing cleat spacing increases cumulative drag resistance, but dense cleats generate cooperative resistance during penetration, diminishing their effective placement and leading to faster yielding of the flowable terrain. These results indicate the existence of an optimal combination of cleat depth and spacing that maximizes locomotor performance without exceeding the robot’s force capacity or compromising effective ground coupling. 

\subsection*{Dual-Plate PIV Analysis}
\begin{figure}
	\centering
	\includegraphics[width=0.99\linewidth]{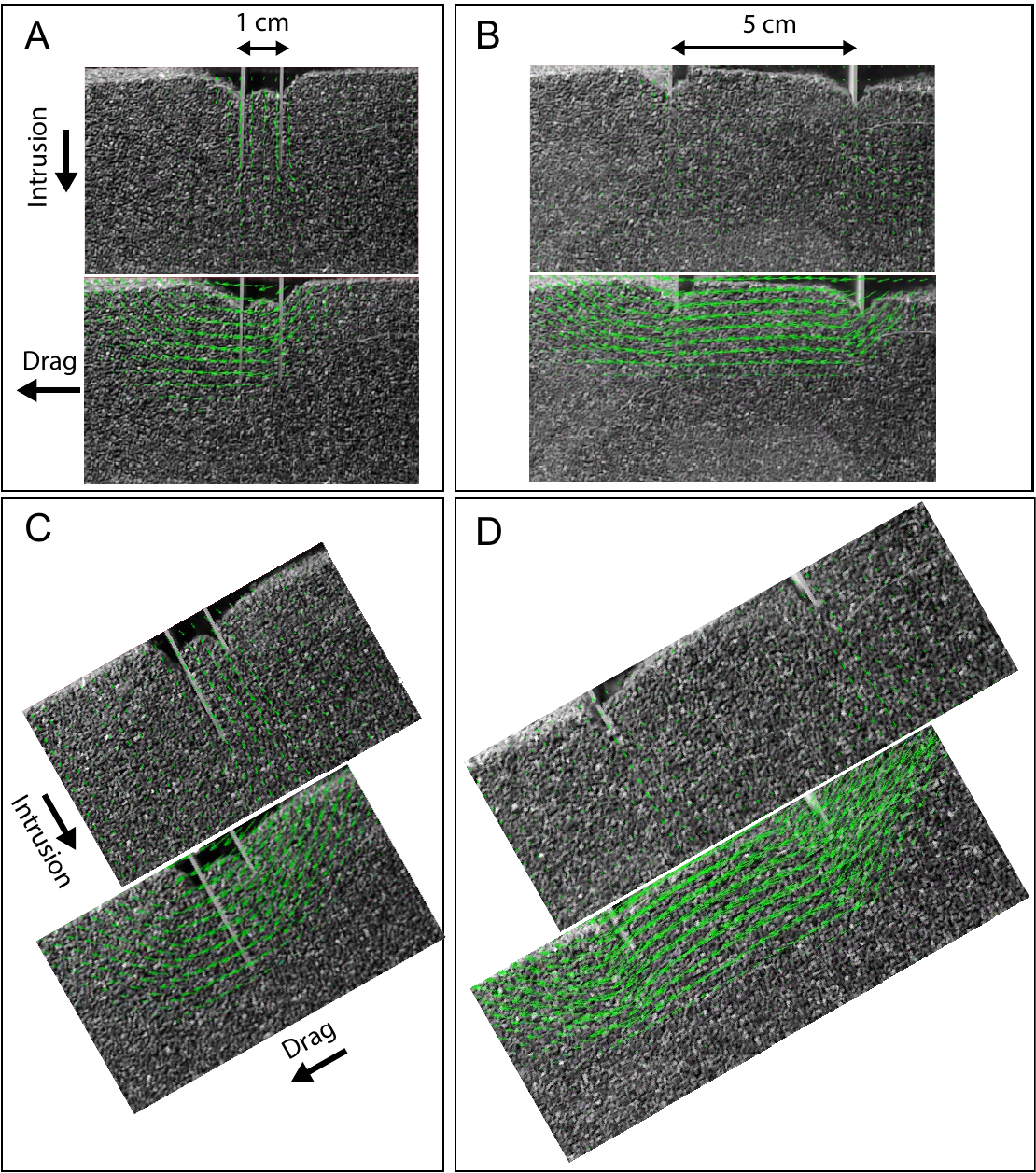}
	\caption{\textbf{Sidewall PIV experiments with dual-plates.} A-B) Dual plates of 1 cm and 5 cm spacing are first inserted and then dragged on level ground, respectively. C-D) Dual plates of 1 cm and 5 cm spacing are inserted and dragged down the slope on a 30 $^\circ$ incline.}
	\label{fig:DualPlatePIV}
\end{figure}

\begin{figure}
    \centering
    \includegraphics[width=.85\linewidth]{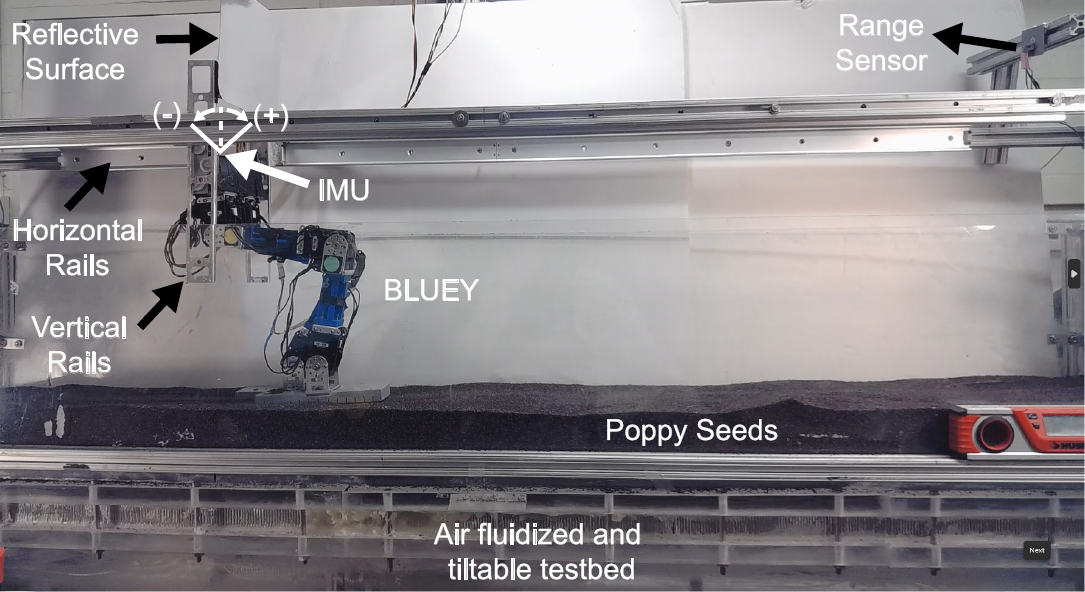}
    \caption{Experimental testbed used to conduct BLUEY locomotion experiments.}
    \label{fig:SI_BlueyTestBed}
\end{figure}

\begin{figure}
    \centering
    \includegraphics[width=.85\linewidth]{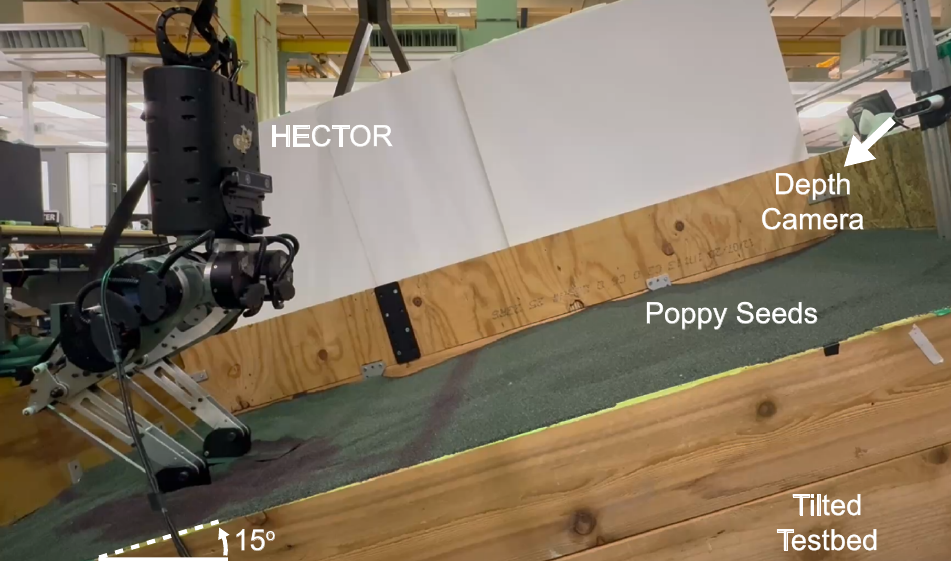}
    \caption{Experimental testbed used to conduct HECTOR locomotion experiments. }
    \label{fig:SI_HectorTestBed}
\end{figure}

\begin{figure}
    \centering
    \includegraphics[width=.85\linewidth]{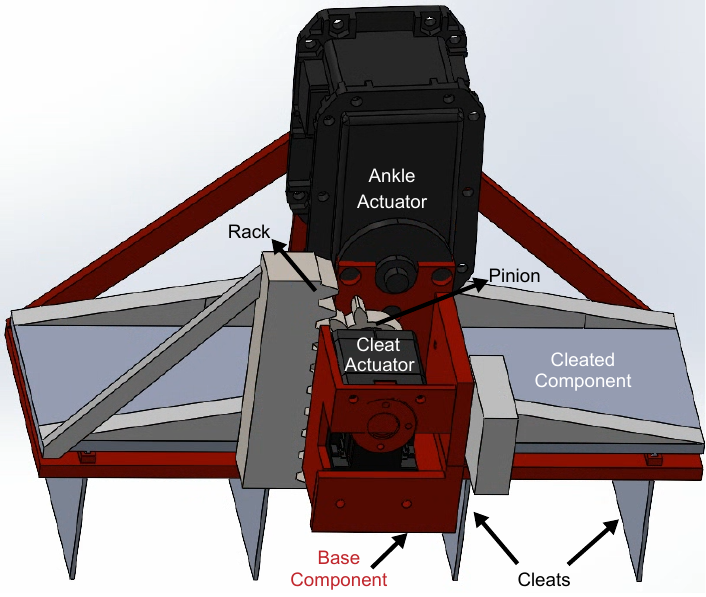}
    \caption{Robotic foot used to perform adaptive cleat extension/retraction experiments using contact sensing. }
    \label{fig:SI_ActCleats}
\end{figure}

\begin{figure}
    \centering
    \includegraphics[width=.9\linewidth]{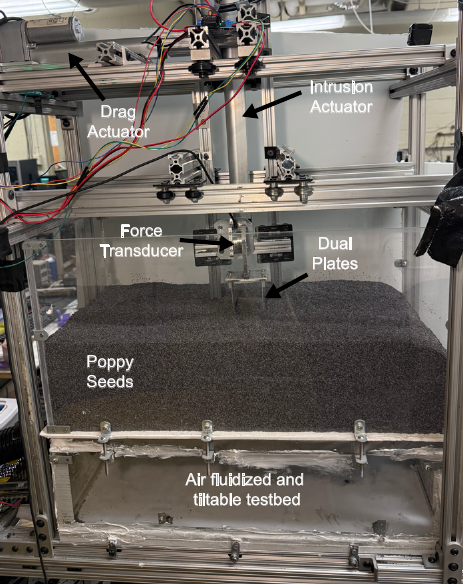}
    \caption{Dual-plate intrusion and drag apparatus mounted on the poppy seed testbed.}
    \label{fig:SI_DualPlates}
\end{figure}

We conducted sidewall PIV experiments using the dual-plate intrusion apparatus to investigate how inter-plate spacing and terrain slope influence the resulting flow fields. The experiments were performed for both 1 cm and 5 cm spaced plate configurations at the level and 30$^\circ$ granular inclines (\refig{fig:DualPlatePIV}). Following the protocol described for the Granular Intrusion Apparatus in the Materials and Methods section, each trial consisted of two phases. First, the plates were inserted into the granular medium to a depth of 3 cm. Subsequently, the plates were dragged downslope. The PIV images were captured at the end of the intrusion and at the onset of plate drag. 

The arrow magnitudes in \refig{fig:DualPlatePIV} represent the bulk particle velocity. The extent of the zones is significantly larger for narrowly spaced cleats than for widely spaced ones, for both level ground and 30$^\circ$ inclines. This increase correlates with the greater volume of material displaced during penetration, increasing penetration resistance. These observations are consistent with prior literature \cite{pravin2021effect,agarwal2021efficacy} and our force measurements. This can be attributed to the overlapping of the flow fields of individual plates. The drag PIV results indicate that the higher drag forces observed for widely spaced plates are likely due to the larger volume of particles trapped between them. Mobilizing these inter-plate particles requires additional force, leading to increased drag resistance for the widely spaced plates, as observed in \refig{fig:CleatForce}. Comparing the drag PIV images of level and inclined terrain for both plate spacings showed that overall particle mobility increased on slopes, while the measured drag forces decreased. This contradiction may be due to the behavior of particles behind the trailing plate. On inclined surfaces, these particles undergo substantially larger displacements than in the level-ground case. However, their motion does not increase drag resistance; instead, it exerts a downward force on the plate, reducing the drag resistance.

\section*{Supplementary Figures}

\section*{Acknowledgments}
This work was supported in part by the National Science Foundation (NSF) under Grant FRR-2328254. Deniz Kerimoglu and Daniel I. Goldman are inventors of a patent application. Helpers: Aman Desai, Burak and Bahadir Catalbas, Edward Kim, Ryan Hirsh, Pengyuan Shu, Cao Ruize.

\bibliography{scibib}

@incollection{atkeson2018happened,
  title = {What happened at the {DARPA} Robotics Challenge finals},
  author = {Atkeson, Christopher G. and Benzun, P. W. and Banerjee, Nandan and Berenson, Dmitry and Bove, Christopher P. and Cui, Xiongyi and DeDonato, Mathew and Du, Ruixiang and Feng, Siyuan and Franklin, Perry and others},
  booktitle = {The {DARPA} Robotics Challenge Finals: Humanoid Robots to the Rescue},
  pages = {667--684},
  year = {2018},
  publisher = {Springer}
}

@article{holmes2006dynamics,
  title = {The dynamics of legged locomotion: Models, analyses, and challenges},
  author = {Holmes, Philip and Full, Robert J. and Koditschek, Daniel E. and Guckenheimer, John},
  journal = {SIAM Review},
  volume = {48},
  number = {2},
  pages = {207--304},
  year = {2006}
}

@article{gu2025robust,
  title = {Robust-Locomotion-by-Logic: Perturbation-resilient bipedal locomotion via signal temporal logic guided model predictive control},
  author = {Gu, Zhaoyuan and Zhao, Yuntian and Chen, Yipu and Guo, Rongming and Leestma, Jennifer K. and Sawicki, Gregory S. and Zhao, Ye},
  journal = {IEEE Transactions on Robotics},
  volume = {41},
  pages = {4300--4321},
  year = {2025},
  doi = {10.1109/TRO.2025.3582820}
}

@article{reher2021dynamic,
  title = {Dynamic walking: Toward agile and efficient bipedal robots},
  author = {Reher, Jenna and Ames, Aaron D.},
  journal = {Annual Review of Control, Robotics, and Autonomous Systems},
  volume = {4},
  number = {1},
  pages = {535--572},
  year = {2021}
}

@techreport{roberts2016rhex,
  title = {{RHex} slips on granular media},
  author = {Roberts, Sonia F. and Koditschek, Daniel E.},
  institution = {University of Pennsylvania},
  type = {Technical Report},
  year = {2016},
  note = {ScholarlyCommons Technical Reports, Department of Electrical and Systems Engineering},
  url = {https://repository.upenn.edu/ese_reports/23}
}

@article{li2013terradynamics,
  title = {A terradynamics of legged locomotion on granular media},
  author = {Li, Chen and Zhang, Tingnan and Goldman, Daniel I.},
  journal = {Science},
  volume = {339},
  number = {6126},
  pages = {1408--1412},
  year = {2013}
}

@article{kamrin2024advances,
  title = {Advances in modeling dense granular media},
  author = {Kamrin, Ken and Hill, Kimberly M. and Goldman, Daniel I. and Andrade, Jose E.},
  journal = {Annual Review of Fluid Mechanics},
  volume = {56},
  number = {1},
  pages = {215--240},
  year = {2024}
}

@article{radosavovic2024real,
  title = {Real-world humanoid locomotion with reinforcement learning},
  author = {Radosavovic, Ilija and Xiao, Tete and Zhang, Bike and Darrell, Trevor and Malik, Jitendra and Sreenath, Koushil},
  journal = {Science Robotics},
  volume = {9},
  number = {89},
  pages = {eadi9579},
  year = {2024}
}

@inproceedings{li2021reinforcement,
  title = {Reinforcement learning for robust parameterized locomotion control of bipedal robots},
  author = {Li, Zhongyu and Cheng, Xuxin and Peng, Xue Bin and Abbeel, Pieter and Levine, Sergey and Berseth, Glen and Sreenath, Koushil},
  booktitle = {Proceedings of the 2021 IEEE International Conference on Robotics and Automation ({ICRA})},
  pages = {2811--2817},
  year = {2021},
  organization = {IEEE}
}

@article{choi2023learning,
  title = {Learning quadrupedal locomotion on deformable terrain},
  author = {Choi, Suyoung and Ji, Gwanghyeon and Park, Jeongsoo and Kim, Hyeongjun and Mun, Juhyeok and Lee, Jeong Hyun and Hwangbo, Jemin},
  journal = {Science Robotics},
  volume = {8},
  number = {74},
  pages = {eade2256},
  year = {2023}
}

@article{westervelt2003hybrid,
  title = {Hybrid zero dynamics of planar biped walkers},
  author = {Westervelt, Eric R. and Grizzle, Jessy W. and Koditschek, Daniel E.},
  journal = {IEEE Transactions on Automatic Control},
  volume = {48},
  number = {1},
  pages = {42--56},
  year = {2003}
}

@article{ames2014human,
  title = {Human-inspired control of bipedal walking robots},
  author = {Ames, Aaron D.},
  journal = {IEEE Transactions on Automatic Control},
  volume = {59},
  number = {5},
  pages = {1115--1130},
  year = {2014}
}

@inproceedings{hereid2014dynamic,
  title = {Dynamic multi-domain bipedal walking with {ATRIAS} through slip-based human-inspired control},
  author = {Hereid, Ayonga and Kolathaya, Shishir and Jones, Mikhail S. and Van Why, Johnathan and Hurst, Jonathan W. and Ames, Aaron D.},
  booktitle = {Proceedings of the 17th International Conference on Hybrid Systems: Computation and Control},
  pages = {263--272},
  year = {2014}
}

@article{mazouchova2010utilization,
  title = {Utilization of granular solidification during terrestrial locomotion of hatchling sea turtles},
  author = {Mazouchova, Nicole and Gravish, Nick and Savu, Andrei and Goldman, Daniel I.},
  journal = {Biology Letters},
  volume = {6},
  number = {3},
  pages = {398--401},
  year = {2010}
}

@article{marvi2014sidewinding,
  title = {Sidewinding with minimal slip: Snake and robot ascent of sandy slopes},
  author = {Marvi, Hamidreza and Gong, Chaohui and Gravish, Nick and Astley, Henry and Travers, Matthew and Hatton, Ross L. and Mendelson III, Joseph R. and Choset, Howie and Hu, David L. and Goldman, Daniel I.},
  journal = {Science},
  volume = {346},
  number = {6206},
  pages = {224--229},
  year = {2014}
}

@article{liao2026failure,
  title = {Failure mechanisms and risk estimation for legged robot locomotion on granular slopes},
  author = {Liao, Xingjue and Qian, Feifei},
  journal = {arXiv preprint arXiv:2603.06928},
  year = {2026}
}

@article{kolvenbach2022traversing,
  title = {Traversing steep and granular {Martian} analog slopes with a dynamic quadrupedal robot},
  author = {Kolvenbach, Hendrik and Arm, Philip and Hampp, Elias and Dietsche, Alexander and Bickel, Valentin and Sun, Benjamin and Meyer, Christoph and Hutter, Marco},
  journal = {Field Robotics},
  volume = {2},
  pages = {910--939},
  year = {2022}
}

@article{yao2024staf,
  title = {{STAF}: Interaction-based design and evaluation of sensorized terrain-adaptive foot for legged robot traversing on soft slopes},
  author = {Yao, Chen and Shi, Guowei and Xu, Peng and Lyu, Shipeng and Qiang, Zhiyang and Zhu, Zheng and Ding, Liang and Jia, Zhenzhong},
  journal = {IEEE/ASME Transactions on Mechatronics},
  volume = {29},
  number = {6},
  pages = {4039--4050},
  year = {2024}
}

@article{shi2024foot,
  title = {Foot vision: A vision-based multifunctional sensorized foot for quadruped robots},
  author = {Shi, Guowei and Yao, Chen and Liu, Xin and Zhao, Yuntian and Zhu, Zheng and Jia, Zhenzhong},
  journal = {IEEE Robotics and Automation Letters},
  volume = {9},
  number = {7},
  pages = {6720--6727},
  year = {2024}
}

@article{godon2024robotic,
  title = {Robotic feet modeled after ungulates improve locomotion on soft wet grounds},
  author = {Godon, S. and Ristolainen, A. and Kruusmaa, M.},
  journal = {Bioinspiration \& Biomimetics},
  volume = {19},
  number = {6},
  pages = {066009},
  year = {2024}
}

@article{piazza2024analytical,
  title = {Analytical model and experimental testing of the {SoftFoot}: An adaptive robot foot for walking over obstacles and irregular terrains},
  author = {Piazza, Cristina and Della Santina, Cosimo and Grioli, Giorgio and Bicchi, Antonio and Catalano, Manuel G.},
  journal = {IEEE Transactions on Robotics},
  volume = {40},
  pages = {3290--3305},
  year = {2024},
  doi = {10.1109/TRO.2024.3415237}
}

@inproceedings{guo2020soft,
  title = {Soft foot sensor design and terrain classification for dynamic legged locomotion},
  author = {Guo, Xiaofeng and Blaise, Bryan and Molnar, Jennifer and Coholich, Jeremiah and Padte, Shantanu and Zhao, Ye and Hammond, Frank L.},
  booktitle = {Proceedings of the 2020 3rd IEEE International Conference on Soft Robotics ({RoboSoft})},
  pages = {550--557},
  year = {2020},
  organization = {IEEE}
}

@article{tyler2023integrating,
  title = {Integrating reconfigurable foot design, multi-modal contact sensing, and terrain classification for bipedal locomotion},
  author = {Tyler, Ted and Malhotra, Vaibhav and Montague, Adam and Zhao, Zhigen and Hammond III, Frank L. and Zhao, Ye},
  journal = {IFAC-PapersOnLine},
  volume = {56},
  number = {3},
  pages = {523--528},
  year = {2023}
}

@article{chen2025dynamic,
  title = {A dynamic foot--terrain interaction model for biped walking on granular media},
  author = {Chen, Xunjie and Huang, Xinyan and Yi, Jingang and Shan, Jerry W. and Liu, Tao},
  journal = {IEEE/ASME Transactions on Mechatronics},
  year = {2025},
  note = {Accepted/In press},
  doi = {10.1109/TMECH.2025.3573687}
}

@inproceedings{chen2024foot,
  title = {Foot shape-dependent resistive force model for bipedal walkers on granular terrains},
  author = {Chen, Xunjie and Anikode, Aditya and Yi, Jingang and Liu, Tao},
  booktitle = {Proceedings of the 2024 IEEE International Conference on Robotics and Automation ({ICRA})},
  pages = {13093--13099},
  year = {2024},
  organization = {IEEE}
}

@inproceedings{xiong2017stability,
  title = {A stability region criterion for flat-footed bipedal walking on deformable granular terrain},
  author = {Xiong, Xiaobin and Ames, Aaron D. and Goldman, Daniel I.},
  booktitle = {Proceedings of the 2017 IEEE/RSJ International Conference on Intelligent Robots and Systems ({IROS})},
  pages = {4552--4559},
  year = {2017},
  organization = {IEEE}
}

@inproceedings{gosyne2018bipedial,
  title = {Bipedal locomotion up sandy slopes: Systematic experiments using zero moment point methods},
  author = {Gosyne, Jonathan R. and Hubicki, Christian M. and Xiong, Xiaobin and Ames, Aaron D. and Goldman, Daniel I.},
  booktitle = {Proceedings of the 2018 IEEE-RAS 18th International Conference on Humanoid Robots ({Humanoids})},
  pages = {994--1001},
  year = {2018},
  organization = {IEEE}
}

@article{karsai2022real,
  title = {Real-time remodeling of granular terrain for robot locomotion},
  author = {Karsai, Andras and Kerimoglu, Deniz and Soto, Daniel and Ha, Sehoon and Zhang, Tingnan and Goldman, Daniel I.},
  journal = {Advanced Intelligent Systems},
  volume = {4},
  number = {12},
  pages = {2200119},
  year = {2022}
}

@article{mikolajczyk2022recent,
  title = {Recent advances in bipedal walking robots: Review of gait, drive, sensors and control systems},
  author = {Mikolajczyk, Tadeusz and Miko{\l}ajewska, Emilia and Al-Shuka, Hayder F. N. and Malinowski, Tomasz and K{\l}odowski, Adam and Pimenov, Danil Yurievich and Paczkowski, Tomasz and Hu, Fuwen and Giasin, Khaled and Miko{\l}ajewski, Dariusz and others},
  journal = {Sensors},
  volume = {22},
  number = {12},
  pages = {4440},
  year = {2022}
}

@article{li2023dynamic,
  title = {Dynamic loco-manipulation on {HECTOR}: Humanoid for enhanced control and open-source research},
  author = {Li, Junheng and Ma, Junchao and Kolt, Omar and Shah, Manas and Nguyen, Quan},
  journal = {arXiv preprint arXiv:2312.11868},
  year = {2023}
}

@inproceedings{kamohara2025rl,
  title = {{RL}-augmented adaptive model predictive control for bipedal locomotion over challenging terrain},
  author = {Kamohara, Junnosuke and Wu, Feiyang and Wamorkar, Chinmayee and Hutchinson, Seth and Zhao, Ye},
  booktitle = {Proceedings of the 2026 IEEE International Conference on Robotics and Automation ({ICRA})},
  year = {2026},
  note = {Accepted; arXiv:2509.18466}
}

@article{pravin2021effect,
  title = {Effect of two parallel intruders on total work during granular penetrations},
  author = {Pravin, Swapnil and Chang, Brian and Han, Endao and London, Lionel and Goldman, Daniel I. and Jaeger, Heinrich M. and Hsieh, S. Tonia},
  journal = {Physical Review E},
  volume = {104},
  number = {2},
  pages = {024902},
  year = {2021}
}

@article{agarwal2021efficacy,
  title = {Efficacy of simple continuum models for diverse granular intrusions},
  author = {Agarwal, Shashank and Karsai, Andras and Goldman, Daniel I. and Kamrin, Ken},
  journal = {Soft Matter},
  volume = {17},
  number = {30},
  pages = {7196--7209},
  year = {2021}
}

@article{gravish2014force,
  title = {Force and flow at the onset of drag in plowed granular media},
  author = {Gravish, Nick and Umbanhowar, Paul B. and Goldman, Daniel I.},
  journal = {Physical Review E},
  volume = {89},
  number = {4},
  pages = {042202},
  year = {2014}
}

@article{mazouchova2013flipper,
  title = {Flipper-driven terrestrial locomotion of a sea turtle-inspired robot},
  author = {Mazouchova, Nicole and Umbanhowar, Paul B. and Goldman, Daniel I.},
  journal = {Bioinspiration \& Biomimetics},
  volume = {8},
  number = {2},
  pages = {026007},
  year = {2013}
}

@article{he2023aerodynamics,
  title = {Aerodynamics and fluid--structure interaction of an airfoil with actively controlled flexible leeward surface},
  author = {He, Xi and Guo, Qinfeng and Xu, Yang and Feng, Lihao and Wang, Jinjun},
  journal = {Journal of Fluid Mechanics},
  volume = {954},
  pages = {A34},
  year = {2023}
}

@inproceedings{kajita2003biped,
  title = {Biped walking pattern generation by using preview control of zero-moment point},
  author = {Kajita, Shuuji and Kanehiro, Fumio and Kaneko, Kenji and Fujiwara, Kiyoshi and Harada, Kensuke and Yokoi, Kazuhito and Hirukawa, Hirohisa},
  booktitle = {Proceedings of the 2003 IEEE International Conference on Robotics and Automation ({ICRA})},
  volume = {2},
  pages = {1620--1626},
  year = {2003},
  organization = {IEEE}
}

@article{askari2016intrusion,
  title = {Intrusion rheology in grains and other flowable materials},
  author = {Askari, Hesam and Kamrin, Ken},
  journal = {Nature Materials},
  volume = {15},
  number = {12},
  pages = {1274--1279},
  year = {2016}
}

\bibliographystyle{Science}

\clearpage

\end{document}